\documentclass[preprint,12pt]{elsarticle}

\usepackage{amssymb}
\usepackage{amsmath,amsfonts,amsthm,bm} 
\usepackage[flushleft]{threeparttable} 
\usepackage{hyperref}
\hypersetup{
 colorlinks = true,
 citecolor= blue,
 linkcolor= blue,
 urlcolor= blue
}
\usepackage[dvipsnames]{xcolor}
\usepackage{subcaption} 
\usepackage{changepage} 
\usepackage{comment}

\journal{Engineering Applications of Artificial Intelligence}

\begin{document}

\begin{frontmatter}

\title{Prediction of wind turbines power with physics-informed neural networks and evidential uncertainty quantification}

\author[inst1,inst2]{Alfonso Gijón\corref{cor1}}
\ead{alfonso.gijon@ugr.es}
\cortext[cor1]{Corresponding author}
\author[inst3]{Ainhoa Pujana-Goitia}
\author[inst3]{Eugenio Perea}
\author[inst1,inst2,inst4]{Miguel Molina-Solana}
\author[inst1,inst2]{Juan Gómez-Romero}

\affiliation[inst1]{organization={Dept. of Computer Science and AI, University of Granada},
    country={Spain}}

\affiliation[inst2]{organization={Research Centre for Information and Communication Technologies (CITIC-UGR), University of Granada},
    country={Spain}}

\affiliation[inst3]{organization={
TECNALIA, Basque Research and Technology Alliance (BRTA)},
    city={Derio},
    country={Spain}}
            
\affiliation[inst4]{organization={Dept. of Computing, Imperial College London},
    country={United Kingdom}}

\begin{abstract}

The ever-growing use of wind energy makes necessary the optimization of turbine operations through pitch angle controllers and their maintenance with early fault detection. It is crucial to have accurate and robust models imitating the behavior of wind turbines, especially to predict the generated power as a function of the wind speed. Existing empirical and physics-based models have limitations in capturing the complex relations between the input variables and the power, aggravated by wind variability. Data-driven methods offer new opportunities to enhance wind turbine modeling of large datasets by improving accuracy and efficiency. In this study, we used physics-informed neural networks to reproduce historical data coming from 4 turbines in a wind farm, while imposing certain physical constraints to the model. The developed models for regression of the power, torque, and power coefficient as output variables showed great accuracy for both real data and physical equations governing the system. Lastly, introducing an efficient evidential layer provided uncertainty estimations of the predictions, proved to be consistent with the absolute error, and made possible the definition of a confidence interval in the power curve.

\end{abstract}

\begin{keyword}
Wind Energy \sep Power Prediction \sep Wind Turbine Modeling \sep Physics-Informed Neural Networks \sep Uncertainty Quantification
\end{keyword}

\end{frontmatter}





\section{Introduction}\label{sec:intro}

The growing use of renewable energies is crucial in addressing climate change and working towards a sustainable energy future. According to the International Renewable Energy Agency (IRENA), annual deployment of around 1000 GW of renewable power is needed to stay on a 1.5\,°C pathway \cite{IRENA}. Offshore wind energy has experimented significant advancements and growth in the last years, establishing itself as a reference technology in the renewable energy sector, and is expected to add up to 160\,GW in Europe by 2030, according to  WindEurope \cite{offshore_growup}. 
At the same time, the thriving development in sensor and storage technology has enabled the collection of ever-increasing amounts of data \cite{Wu2016,AKHAVANHEJAZI2018}. This, combined with the development of flexible and powerful data-driven and machine-learning approaches, introduces a new paradigm of digital, autonomous, and decentralized control for manufacturing systems \cite{SAHAL2020,Gomez2015,Canizo2017}. In this scenario, it is crucial to have accurate and robust wind turbine (WT) models for operations optimization and automatic diagnoses of faults.

Regarding wind energy, current devices paired with advanced analytic capability enable real-time performance monitoring, optimization of turbine operations, and predictive maintenance through early fault detection  \cite{ML-WindTurbines-Review}, thus minimizing downtime \cite{offshore_digitalization}. However, data-based WT modeling can present several challenges \cite{WT_modelling_challenges}, making complex the direct application of data-driven techniques to the field. The first challenge is obtaining data in sufficient quantity and quality on various parameters, such as wind speed and direction, power generated, or performance data.
Another difficulty arises from the complexity of WTs, involving multiple components and interrelated variables. 
Furthermore, wind variability is another obstacle, as it is a highly fluctuating source of energy 
which sometimes requires complex techniques to mitigate the operational distortion \cite{AVENDANOVALENCIA2020106686}. Traditional physics-based models are becoming insufficient when faced with the system complexity and its various uncertainties, so validating and calibrating a WT model from measured data is vital to ensure its accuracy and reliability \cite{en16020861}. 
Overcoming these challenges requires a combination of technical knowledge, experience in the field, and access to relevant and reliable data.

Power prediction in WTs plays a fundamental role in wind energy management and power forecasting \cite{WT_power_curve}. In addition to its applications in the management of generated energy and integration into the electricity grid \cite{10.1049/ip-gtd_19941215}, power prediction combined with data analysis and monitoring techniques can help to detect and diagnose powertrain component failures \cite{en14185967}. By comparing power predictions with real-time data, it is possible to identify deviations and anomalies that could indicate component problems, enabling early intervention and effective fault management. Besides, power prediction is essential for effective pitch angle control and optimal WT performance, as it provides real-time information about the power generated by the WT. This information can also be used to adjust blade angles to maximize power generation, protect the WT, and achieve efficient overall control \cite{APATA2020e00566}. 

The lack of precise and robust physics-based models for predicting the power production of utility-scale wind farms \cite{Howland2019} incentives the application of data-driven approaches. Moreover, as wind-power data are often very noisy, fitted WT power curves could differ significantly from the theoretical ones provided by manufacturers \cite{WT_power_curve}. It is well known the success of deep learning methods (especially artificial neural networks) for multivariate regression tasks in a purely data-driven way, that is, when there is a sufficient amount of data available, but the physical picture is partially unknown or too complex. Although neural networks are black-box models, the development of new architectures able to respect certain constraints, like physics-informed neural networks (PINNs) \cite{RAISSI2019,FMata2023}, makes them more attractive for modeling physical phenomena. PINNs can reproduce real data while respecting some physical laws, giving rise to accurate and robust models. Based on those ideas, this work presents the design of several wind-power models that achieve great accuracy while fulfilling some fundamental physical equations governing the system. 

Furthermore, uncertainty quantification on deep learning models could be one of the main strengths of neural networks as universal regressors compared to standard methods (usually based on the optimization of a set of parameters from analytical formulas constructed from first principles in the best case, but usually based on physics intuition or empirical information) to reproduce real data. Reliability and uncertainty modeling was described as a key research challenge for wind energy by the European Academy of Wind Energy (EAWE) \cite{Kuik2016}. In our work, uncertainty is easily estimated through an evidential deep learning layer \cite{evidential2020}, which allows us to obtain a power curve model consistent with the specifications provided by the manufacturer. 

The main goal of this work is to design and validate robust and accurate prediction models for the WT power, which is the previous step to developing optimal controllers and applying failure detection methods. We built neural network models not only for the power coefficient but also for the torque and the power itself, comparing their performances in predicting the power generated by the turbine. Further, we introduced the use of physics-informed neural networks to model those magnitudes. 
For each model, we added an evidential layer to the output, a recent and efficient method to quantify the uncertainty of the prediction, which is key to identify data deviated from normal. At the end, we have accurate models for prediction of the WT variables, including the generated power, available at the repository \href{https://github.com/alfonsogijon/WindTurbines_PINNs}{https://github.com/alfonsogijon/WindTurbines\_PINNs}\footnote{All the scripts for data preprocessing, building, training and evaluating the models are available at this github repository, together with the reduced dataset without the non-physical and anomalous data points. This allows the reproduction of all the figures and results presented in the paper.}. Even though these models have been trained with a dataset of a specific manufacturer, they could be easily transferred to different manufacturing data.

The rest of the paper is organized as follows. Section \ref{sec:background} introduces the physical background of the problem and provides context discussing some related work. Section \ref{sec:methods} 
establishes the experimental setup and presents the methods employed to build the WT models. Data explanation and the discussion of the experiments are found in Sections \ref{sec:data} and \ref{sec:experiments}, respectively. Finally, Section \ref{sec:conclusions} summarizes the main conclusions of this work.

\section{Background}\label{sec:background}

\subsection{Physics of wind turbines}

The low-scale behavior of a wing turbine can be modeled with physics-based approaches. However, that is very challenging due to the complexity of fluid flows that must consider several aspects such as geophysical and atmospheric effects, the turbulence at high Reynolds numbers, as well as wind-turbine characteristics and wind-farm layouts, among others \cite{10.1063/5.0091980}. While low-order, physics-based models are helpful for qualitative physical understanding, they generally are unable to accurately predict the power production of utility-scale wind farms due to a large number of simplifying assumptions and neglected physics \cite{Howland2019}.

\begin{figure}[t]
\begin{subfigure}{0.55\textwidth}
\includegraphics[width=1.0\linewidth]{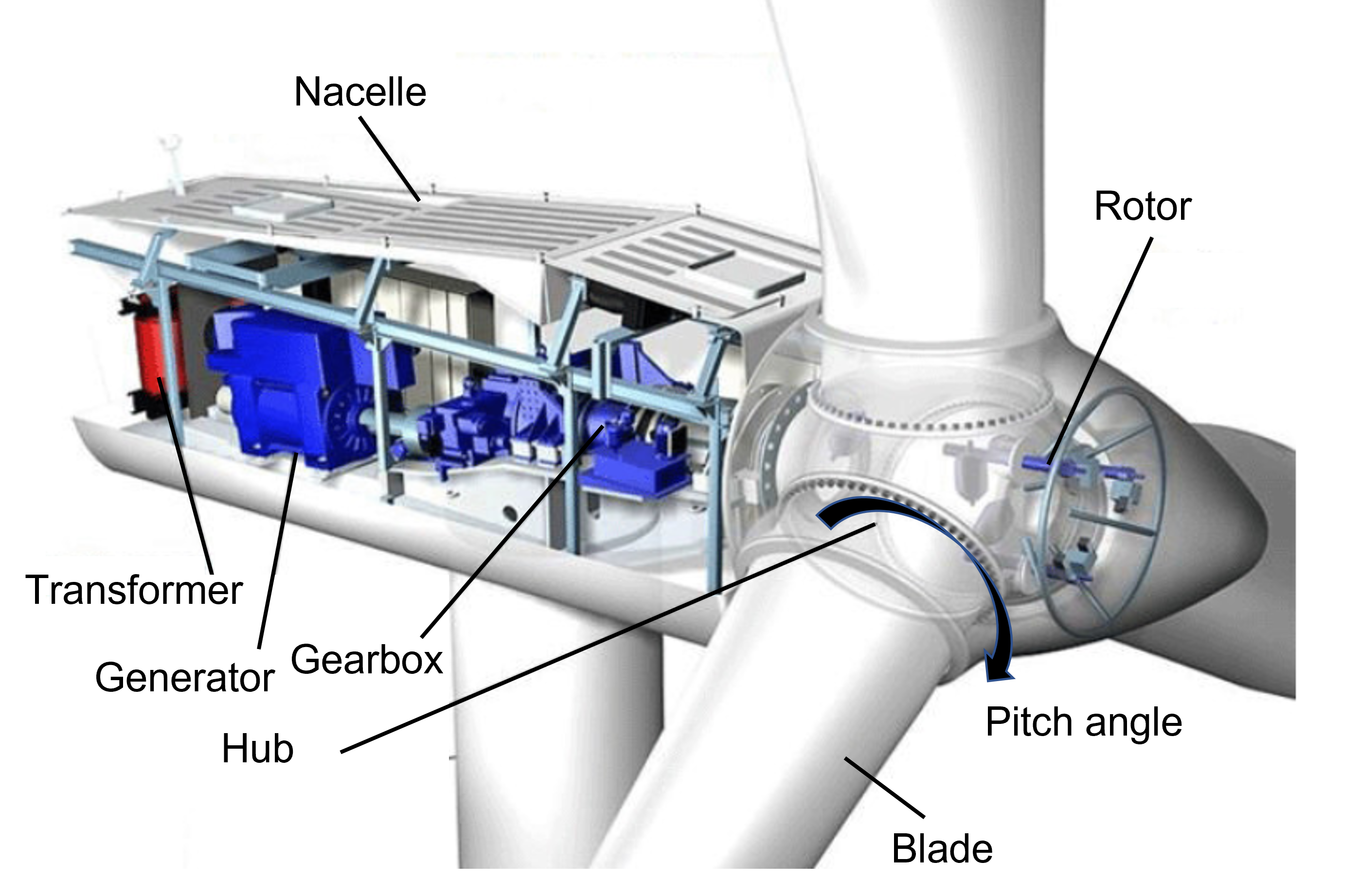} 
\caption{}
\label{fig:scheme1}
\end{subfigure}
\begin{subfigure}{0.55\textwidth}
\includegraphics[width=1.0\linewidth]{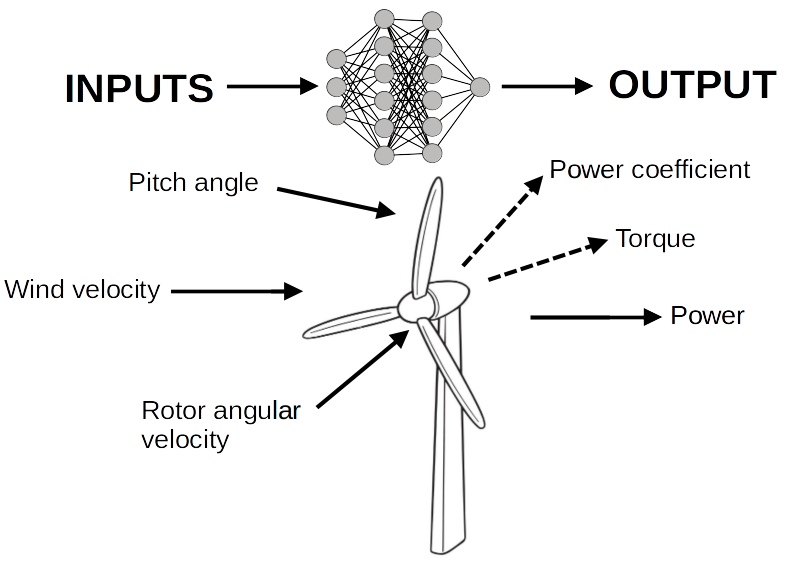}
\caption{}
\label{fig:scheme2}
\end{subfigure}

\caption{(a) Main components of a wind turbine. Image adapted from \href{https://www.linkedin.com/pulse/smart-grid-energy-harvesting-martin-ma-mba-med-gdm-scpm-pmp/}{Smart Grid: Energy Harvesting - LinkedIn}  (b) Scheme of a neural network model, receiving the pitch angle, wind velocity and rotor angular velocity as input arguments and predicting power coefficient, torque or power as output.}
\label{fig:schemes}
\end{figure}

However, partial physical information is available through well-established equations relating some high-scale variables. From a mechanical point of view, the generated power $P$ is given by
\begin{equation}\label{eq:Power-torque}
    P = g T \omega \,,
\end{equation}
where $g$ is the drivetrain gear ratio (a constant property characteristic of the turbine model), $T$ is the mechanical torque of the rotor, and $\omega$ is its angular velocity.
 This equation can be interpreted as a state equation of a WT, or a conservation/symmetry law that must be fulfilled independently of the underlying wind-power model. In turn, the power extracted by a WT from the kinetic energy of the incoming wind yields:
\begin{equation}\label{eq:Power-cp}
    P = \frac{1}{2} C_{p} \rho A v^3\,,
\end{equation}
where $C_p$ is the power coefficient, $\rho$ is the air density, $A$ is the swept area by the blades of the WT and $v$ is the wind velocity. The physical parameters of the extended turbine model for the MM82 case are shown in \autoref{tab:MM82-parameters}, while a scheme of the elements of a WT can be seen in \autoref{fig:scheme1}. 


\begin{table}
\renewcommand{\arraystretch}{1.2}
\centering
\begin{tabular}{|c|c|c|c|c|}
\cline{2-5} \cline{3-5} \cline{4-5} \cline{5-5} 
\multicolumn{1}{c|}{} & $R\,$(m)  & $A\,\text{{(m\texttwosuperior)}}$ & $g$ & $\rho\,\,\text{{(kg/m\textthreesuperior)}}$\tabularnewline
\hline 
MM82 & 41 & 5281 & 106 & 1.225\tabularnewline
\hline 
\end{tabular}
\caption{Physical parameters of the MM82 wind turbines, obtained from \cite{MM82}.}
\label{tab:MM82-parameters}
\end{table}

The power generated by a WT strongly depends on the power coefficient, $C_p$, a dimensionless quantity influenced by the intrinsic characteristics of a WT (size, geometry, aerodynamic properties), and the operating conditions, defined by the wind speed and pitch angle variables. Usually, the objective is to maximize the value of $C_p$ to achieve the highest efficiency in the conversion of wind energy into electrical energy \cite{doi:https://doi.org/10.1002/9781119992714.ch3,IEC61400-12-1}, subjected to a theoretical maximum value of 0.5926 (the Betz limit). The maximum power of WTs in variable wind speed regions is achieved by a suitable control of the pitch angle $\beta$, defined as the angle between the lateral axis of the blades and the relative wind. Pitch control of WTs is complex due to the intrinsic non-linear behavior of these systems and external disturbances, due to changing wind conditions and other meteorological phenomena. Therefore, an accurate estimation of the $C_p$ coefficient and/or the generated power as a function of the wind speed and the operating conditions is essential for effective pitch control and fault detection. 


\subsection{Modeling of the power}

It is important to note that the power coefficient is not constant and can vary with changes in wind speed, turbulence, and other operating conditions, so that different WT designs and operating strategies may have different $C_p$ characteristics. Therefore, due to its complexity, $C_p$ is often analyzed and optimized empirically during the design, operation, and control of WTs to maximize their energy conversion efficiency. However, WTs tipically operates in complex and changing conditions, which are far from the stable experimental conditions, so the information provided by manufacturers does not usually accurately reflect the actual performance of WTs in real-world scenarios \cite{PAGNINI2015,YAN2019}.

Though the exact value of the power coefficient $C_p$ is determined by complex underlying fluid dynamics processes, it can also be approximated as a function of the high-scale variables. In this way, the conversion efficiency of a WT depends on the input variables $(v,\beta,\omega)$ and different empirical formulas have been proposed for the power coefficient  \cite{Jamdade2015,Carpintero2020,cp_equations}, being the exponential one of the most widely used: 
\begin{equation}\label{eq:Cp-empirical}
    C_p(\lambda,\beta) = c_0 (c_1\gamma - c_2\beta - c_3\beta^{c_4} -c_5)e^{-c_6\gamma} + c_7 \lambda \,,
\end{equation}
where $\lambda=\omega R/v$ represents the tip speed ratio, $\beta$ is the pitch angle and $\gamma$ is defined by 
\begin{equation}
    \gamma = \frac{1}{\lambda+d_0\beta+d_1} - \frac{d_2}{1+\beta^3} \,.
\end{equation}
This empirical model has eleven free parameters, denoted as ${c_i}$ and ${d_i}$, fitted to reproduce real measurements. The main advantage of parametric models is their simple form, but sometimes is hard to accurately fit the complex relationship between wind speed and power. In contrast, non-parametric methods are more accurate and do not require an explicit expression to learn automatically from large amounts of data. However, they can suffer overfitting, significant discrepancies with the manufacturer specifications, and other problems when trained on real datasets \cite{WANG2023}.

\subsection{Related work}\label{subsec:related-work}

In recent years, intelligent techniques have been employed for pitch angle control and fault detection \cite{Zhang2008,Abdul-Ruhman2020,SierraGarcia2022-2,Karami2023, Bilendo2021, SUN2021}, trying to maximize the generated power. These studies usually use a parametric empirical formula for the power coefficient $C_p$, with exponential, polynomial or sinusoidal form, where the free parameters are fitted to approximately reproduce the data at hand \cite{Jamdade2015,Carpintero2020,cp_equations}. Recently, some authors have proposed using neural network models for regression of the power coefficient, improving the quality of the prediction of the generated power \cite{WANG2023,MANOBEL2018,LI2022}. The accurate mathematical models of the system provided by neural networks open the door to applying powerful control methods such as deep reinforcement learning \cite{SierraGarcia2021,SierraGarcia2022,Coquelet2022, XIE2023}, which require the exploration of many configurations of the environment. The direct application of data-driven and, especially, deep learning methods can lead to good results in predicting the WTs power curve and play a crucial role in turbine performance analysis, but they suffer a lack of robustness and physical sense. To find an optimal controller, the predictive model should robustly reproduce real data, respecting the physical restrictions imposed by the system even for configurations that deviate from normal behavior. Some authors have used physics-informed machine learning to build wake models representing the flow field behind a WT
\cite{Howland2019} or for fatigue damage prediction \cite{LI2022-2}. However, as far as we know, PINNs have never been applied to respect the high-scale global equations relating the outputs (power coefficient, torque, and power) with the input variables (wind speed, pitch angle, and angular velocity), considered in this work as conservation or symmetry laws, without imposing further internal restrictions in the form of the predicted quantities. 

Moreover, uncertainty modeling of WTs is still in its infancy but constitutes a fundamental aspect for minimizing unexpected failures and downtime \cite{Kuik2016}. Some Bayesian \cite{Jacques2023,Aerts2023} or Monte Carlo \cite{Marepally2022,Pascal2022,Perr-Sauer2021} methods have been recently employed to approximate the uncertainty in estimating the WT power curve and wake models, though they are computationally expensive. In our work, we equipped the neural network models for power prediction with a final evidential layer \cite{evidential2020}, an efficient and easily implemented feature to quantify epistemic uncertainty, making the model able to accurately reproduce both real data and the manufacturer specifications within the confident interval.

\section{Methods}\label{sec:methods}

\subsection{Problem statement} 
This work aims to predict the power $P$ generated by WTs as a function of the high-scale input variables $(v,\beta,\omega)$. 
Each WT configuration is defined by the 5-tuple $(v,\beta,\omega,T,P)$, whose values are measured on real operation and collected into datasets. The power coefficient $C_p$ can be computed straightforwardly from the generated power and wind velocity, to measure the energy efficiency of each WT configuration.
In addition to the data, physical information is given by Equations \ref{eq:Power-torque} and \ref{eq:Power-cp}, providing algebraic relations between the state variables. Note that although the magnitude of interest is the generated power, it could be easily obtained from intermediate magnitudes like the torque or the power coefficient using those physical equations. 

To solve the regression task of this study, a data-driven approach is proposed, able to learn directly from real data without the need for empirical parametrizations. Specifically, neural networks were used to learn the relation between the output and the input variables of the WT, as represented in \autoref{fig:scheme2}. The accuracy of models with different target magnitudes, such as power coefficient, torque, and power, can be compared by analyzing the error metrics on the predicted power. Physical equations were imposed on the training to increase the robustness of our data-driven models, and an uncertainty quantification mechanism was implemented.

\subsection{Experimental setup}

Our experimental setup is based on artificial neural network models, trained using the \href{https://www.tensorflow.org/}{\emph{Tensorflow}} library \cite{tensorflow} and hyperparameters optimized with the Hyperband algorithm \cite{hyperband} from the \href{https://keras.io/keras_tuner/}{\emph{Keras Tuner}} library \cite{keras}. PINNs were implemented with the \href{https://www.sciann.com/}{\emph{SciANN}} library \cite{sciann}, using optimal architectures previously found for standard artificial neural networks. For uncertainty quantification, evidential layers were added using the \href{https://github.com/aamini/evidential-deep-learning}{\emph{Evidential Deep Learning}} library \cite{evidential2020}. All the calculations were carried out with a desktop computer with an 11th Gen Intel Core i7-11800H processor, 16 GB RAM memory and NVIDIA GeForce RTX 3060 graphic card.

\subsection{Physics-Informed Neural Networks}
As commented previously, a PINN model can learn from data and also from equations, which can be totally known or dependent on some unknown parameters. In the latter case, the training of the model results in the learning of both the desired quantity and the model parameters that best describe the observed data. In time-independent systems, a PINN is just an artificial neural network, $\text{NN}(x;W,b)$, trying to approximate a target function $u(x)$, both differentiable functions of the input $x$. Sometimes, $u$ is known to fulfill some physical conditions or constraints, which generally can be expressed as ordinary or partial differential equations, written as
\begin{equation}
    F \big[ u(x);\mu \big] = 0 \,,
\end{equation}
where $F$ is a functional on the solution function $u$ and the model parameters $\mu$. The network can learn the target function and the optimal parameters of the physical equation by minimizing the loss function\footnote{Although the loss is expressed here as the mean absolute error between prediction and target, all the arguments and conclusions are valid for different loss functions.} with respect to the weights and biases of the neural network and the model parameters: 
\begin{subequations}
\begin{align}
   L & = L_{\text{data}} + L_{\text{phys}} \,, \\
   L_{\text{data}} & = \frac{1}{N_\text{data}}\sum_{i=1}^{N_\text{data}} \big| \text{NN}_i - u_i\big| \,, \\ 
   L_{\text{phys}} & = \frac{1}{N_\text{phys}}\sum_{j=1}^{N_\text{phys}} \big| F\big[\text{NN}_j;\mu \big] \big| \,,
\end{align}
\end{subequations}
where $\text{NN}_i=\text{NN}(x_i;W,b)$ is the neural network model evaluated at $x_i$. The physical term can be considered as a regularization term to the effect of tuning the weights of the neural network to respect a physical condition expressed by equation $F$. 

In our case, the target function $u$ has to fulfill physical ordinary equations given by \ref{eq:Power-torque} and \ref{eq:Power-cp}, which directly provides an estimation of the target magnitude, $\Tilde{u}$, as a function of the inputs. Hence, considering there are no physical parameters to be optimized, the functional $F$ has the simple form \begin{equation}
    F[u(x)] = u(x) - \Tilde{u}(x) \,.
\end{equation}

First, one could consider the power as the target function, $u\equiv P$. In this case, the inputs of the PINN are $(v,\beta,\omega,T)$, though only $(v,\beta,\omega)$ are inputs of the neural network, and $T$ is only used into the physical term. Details about the model architecture can be consulted in \ref{sec:PINN_arhitecture}. In addition to the regression of the $P$ data, an estimation of the power can be computed from \autoref{eq:Power-torque}, $\Tilde{u}=gT\omega$, so the physical condition is expressed as
\begin{equation}
   \text{NN}(v,\beta,w) - gT\omega = 0 \,,
\end{equation}
while the residuals are:
\begin{subequations}
\begin{align}
   \label{eq:Ldata_P}
   L_{\text{data}} & = \frac{1}{N}\sum_{i=1}^{N} \big| \text{NN}_i - P_i \big| \,, \\ 
   \label{eq:Lphys_P}
   L_{\text{phys}} & = \frac{1}{N}\sum_{i=1}^{N} \big| \text{NN}_i - gT_i\omega_i \big| \,.
\end{align}
\end{subequations}
The training of this PINN results in a neural network predicting the measured power while respecting the mechanical relation between power and torque. 

Secondly, the power coefficient factor could be considered as the target function of a PINN, $u\equiv C_p$. Similarly to the previous case, the inputs of the PINN are $(v,\beta,\omega,T)$, but only $(v,\beta,\omega)$ are inputs of the neural network, and $T$ is only used for the physical residual. In this case, an estimation of the power coefficient can be computed by equaling the right-hand sides of Equations \ref{eq:Power-torque} and \ref{eq:Power-cp} to obtain $\Tilde{u}(v,T,\omega)$, so the imposed condition is
\begin{equation}
   \text{NN}(v,\beta,w) - \frac{2gT\omega}{\rho A v^3} = 0 \,,
\end{equation}
and the residuals are:
\begin{subequations}
\begin{align}
   L_{\text{data}} & = \frac{1}{N}\sum_{i=1}^{N} \big| \text{NN}_i - C_{p,i}\big| \,, \\ 
   L_{\text{phys}} & = \frac{1}{N}\sum_{i=1}^{N} \big| \text{NN}_i - \frac{2gT_i\omega_i}{\rho A v_i^3} \big| \,.
\end{align}
\end{subequations}
The training of this PINN results in a neural network predicting the power coefficient while respecting physical relations. The generated power is computed from the predicted $C_p$ by means of \autoref{eq:Power-cp}.

Lastly, a PINN with the torque $T$ as target function would lead to the equation
\begin{equation}
   \text{NN}(v,\beta,w) - \frac{P}{g\omega} = 0 \,.
\end{equation}
In this case, the inputs of the PINN are $(v,\beta,\omega,P)$, but only $(v,\beta,\omega)$ are inputs of the neural network and $P$ is only used to impose the physical equation. Residuals are:
\begin{subequations}
\begin{align}
   \label{eq:Ldata_T}
   L_{\text{data}} & = \frac{1}{N}\sum_{i=1}^{N} \big| \text{NN}_i - T_i\big| \,, \\ 
   \label{eq:Lphys_T}
   L_{\text{phys}} & = \frac{1}{N}\sum_{i=1}^{N} \big| \text{NN}_i - \frac{P_i}{g\omega_i} \big| \,.
\end{align}
\end{subequations}

It should be noticed that torque is just an intermediate quantity for computing the interest output, the power, through the equation $P_T=gT\omega$, so that the physical condition is satisfied by construction. The residual $L_{\text{data}}$ from \autoref{eq:Ldata_T} contributes to the regression of the torque data and hence, a power coming from the formula $P_T=gT\omega$, fulfilling \autoref{eq:Power-torque}. In turn, the introduction of the variable $P$ into the residual $L_{\text{phys}}$ of \autoref{eq:Lphys_T} leads to a better prediction of the measured power $P$, because it leads the model to learn $T_P=P/g\omega$. Paradoxically, due to the construction of the PINN for the regression of the torque and the subsequent calculation of the power from it, $L_{\text{data}}$ tends to fulfill the physical condition whereas $L_{\text{phys}}$ is intended to the regression of the measured power.




\subsection{Evidential layer for uncertainty quantification}
Uncertainties were evaluated by including an evidential deep learning layer to the output \cite{evidential2020}. While representations of aleatoric uncertainty can be learned directly from data, several approaches exist for estimating epistemic uncertainty. Evidential deep learning assumes that observed targets are drawn from a Gaussian distribution with unknown mean and variance, $y\sim\mathcal{N}(\mu,\sigma^2)$. Accordingly, the mean and variance are represented as
\begin{subequations}
\begin{align}
   \mu & \sim \mathcal{N}(\gamma,\sigma^2,\nu^-1) \,, \\[2mm]
   \sigma^2 & \sim \Gamma^{-1} (\alpha,\beta) \,,
\end{align}
\end{subequations}
where $\Gamma$ is the gamma function and $(\gamma,\nu,\alpha,\beta)$ are parameters. The posterior distribution follows a normal inverse gamma distribution from which the prediction and epistemic uncertainty are computed from the following equations, respectively:
\begin{subequations}
\begin{align}
   \mathbb{E}(\mu) & = \gamma \,, \\[2mm]
   \text{Var}(\mu) & = \frac{\beta}{\nu(\alpha-1)} \,.
\end{align}
\end{subequations}
The loss function for training the evidential deep learning model includes a negative likelihood loss $L^{\text{NLL}}$ that is responsible for maximizing the model prediction, and an evidential loss $L^{\text{EL}}$ which minimizes the evidence on errors. The total loss consists of the two terms for maximizing prediction and regularizing evidence, scaled by a coefficient $\lambda$:
\begin{equation}
    L(x) = L^{\text{NLL}}(x) + \lambda L^{\text{EL}}(x) \,.
\end{equation}
Here, $\lambda$ trades off uncertainty inflation with model fit. Setting $\lambda$=0 yields an over-confident estimate while setting $\lambda$ too high results in over-inflation. The hyperparameter $\lambda$ was set to $0.01$ within this work, a good choice for similar regression problems \cite{evidential2020}.

\section{Data}\label{sec:data}
Real data, obtained from the website of \href{https://opendata-renewables.engie.com/}{ENGIE Renewables}, were collected from 4 different MM82 turbines from the Senvion manufacturer at `La Haute Borne' wind farm \cite{LaHauteBorne1,LaHauteBorne2}, into two runs between 2013-2016 and 2017-2020. Among the available variables related to the physical conditions and the turbine itself, we focus our study on wind velocity, pitch angle, rotor speed, torque, and power, represented as $(v,\beta,w,T,P)$. Our dataset is composed of 1057968 5-dimensional vectors. A value of the power factor $C_p$ can be computed for each data point by inverting \autoref{eq:Power-cp} and using measured $P$ and $v$ as input variables. Data were obtained from direct or indirect measurement, so they contain noise; thus, some preprocessing is desirable.

\begin{figure}[t]
    \centering
    \includegraphics[width=1.0\linewidth]{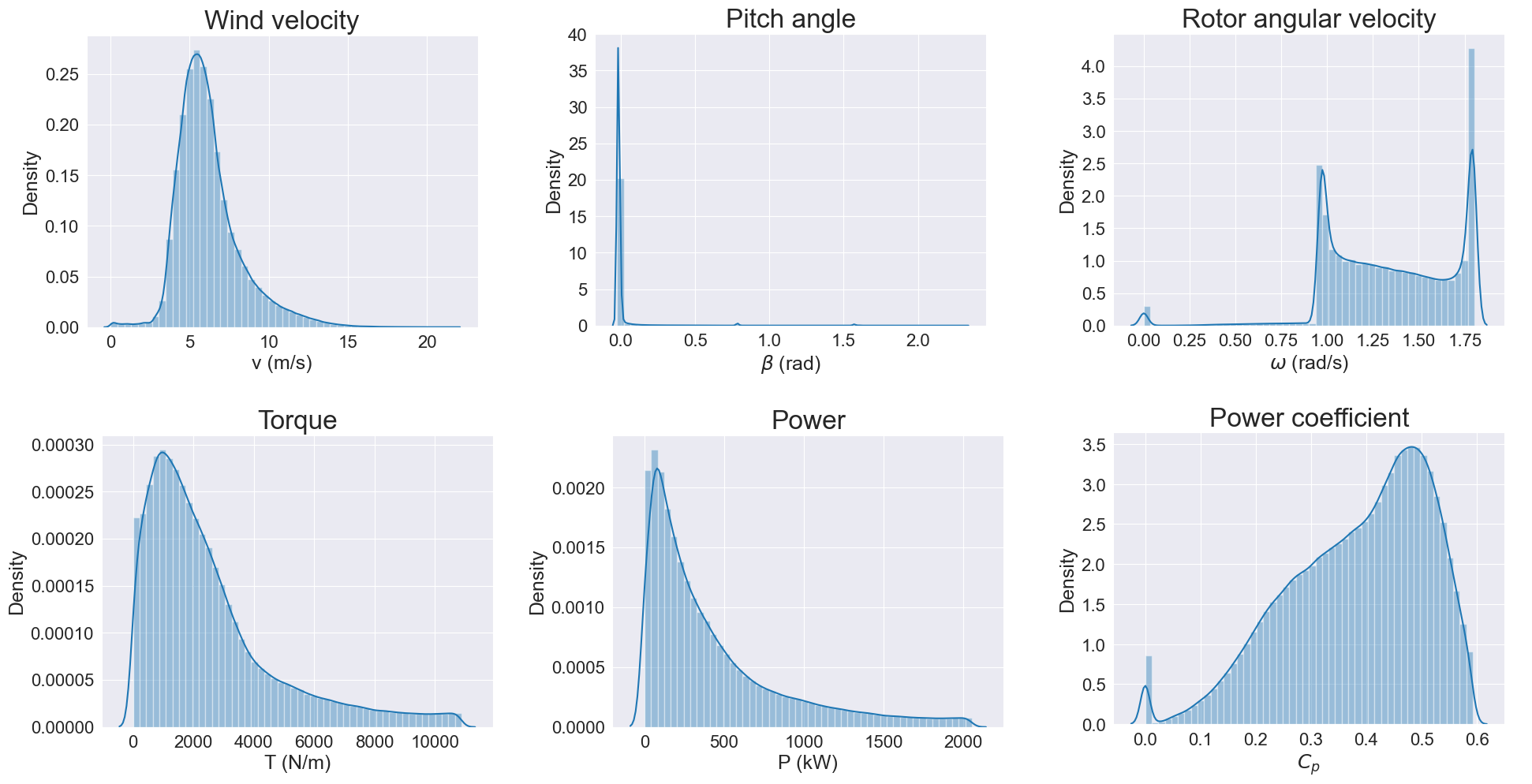}
    \caption{Distribution of the variables $(v,\beta,\omega,T,P,C_p)$ on the dataset.}
    \label{fig:VarDist}
\end{figure}

\begin{figure}[t]
    \centering
    \includegraphics[scale=0.8]{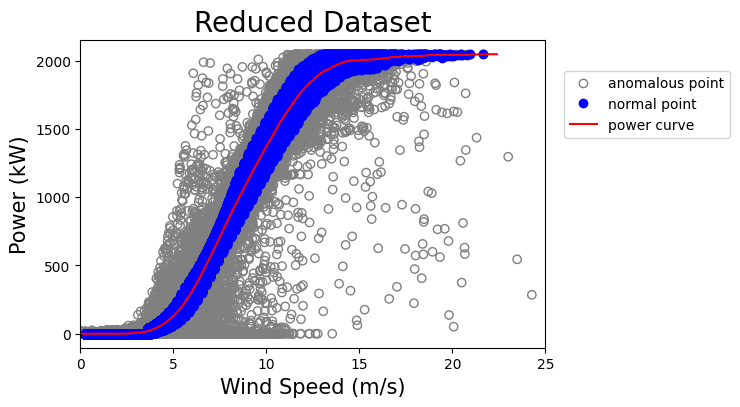}
    \caption{Estimation of the power curve (red) of the turbines of our dataset, along with normal (blue) and anomalous (grey) measured data-points.}
    \label{fig:Power_curve_data}
\end{figure}

First, we removed all non-physical data, e.g., those having a power coefficient higher than the theoretical Betz limit, $C_p>0.5926$. After that, we identified anomalous data by comparing measured power with an estimation of the power curve based on an iterative median technique \cite{Minh2018}. Each data having a power further than three sigma from the median was discarded. Low-velocity data are especially noisy, producing physical but unreal values of the power coefficient $C_p$, obtained by dividing by $v^3$; therefore, we applied a power cutoff of $25\,$kW for velocities lower than $3.5\,$m/s, and data with a higher power were removed from the dataset. Actually, that power cutoff at that threshold velocity is the lowest power value provided by the manufacturer for MM82 turbines \cite{MM82}. The distribution of the variables can be seen in \autoref{fig:VarDist}, while the estimated power curve along with the discarded data is represented in \autoref{fig:Power_curve_data}.

As a result of preprocessing, the dataset was reduced from more than 1 million data to $7.27\times10^5$ data, approximately a $70\,\%$ of the original dataset. The preprocessing allowed us to avoid anomalous data, coming from failure or a non-optimal behavior, which could difficult the learning of our models, but it did not remove the inherent noise of the data. Such noise can be quantified by comparing the power data itself, $P$, with the power estimated from the mechanical relation of \autoref{eq:Power-torque}, using the torque and rotor angular velocity data, $P_T=gT\omega$. On average, we obtain a mean absolute percentage error of $1.7\,\%$, which establishes a lower limit for any model trying to predict the power by taking noisy data as input. 

\section{Experimentation and discussion}\label{sec:experiments}

\begin{figure}
    \centering
    \includegraphics[width=0.7\linewidth]{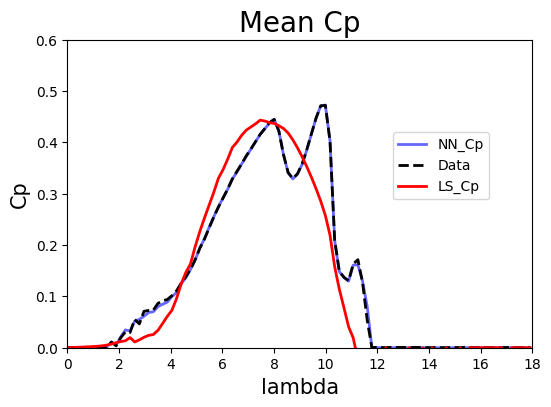}
    \caption{Comparison of mean $C_p$ curves obtained from data (black dash line), a neural network model (blue) and the parametric model from the empirical \autoref{eq:Cp-empirical} with optimal parameters found with the least squares method (red).}
    \label{fig:Cp_curve}
\end{figure}

The first attempt in building a model for the generated power is using the parametric model given by the empirical \autoref{eq:Cp-empirical}. The parameters that best describe the $C_p$ data can be found using the non-linear least squares method. However, this fitting was not satisfactory because the employed parametrization is not general enough to reproduce the data, as can be seen in \autoref{fig:Cp_curve}, where the mean $C_p$ curve obtained from data (dash and black line) is compared with the optimal empirical fit (red line). The fitting drastically improves when an artificial neural network is used as a regression model, being able to capture different regimes in the curve. In that case, the predicted $C_p$ curve, in blue, perfectly matches the target one, dash black line. Performance metrics\footnote{The accuracy of different regression models of the power are compared through commonly used metrics, such as mean absolute error (MAE), root mean squared error (RMSE), mean absolute percentage error (MAPE) and R-squared (R2).} of these two fits can be found in \autoref{tab:metrics_CP}. As expected, the neural network model vastly outperforms (by one order of magnitude for the error metrics) the empirical model. 

\begin{table}
    \centering
    \renewcommand{\arraystretch}{1.2} 
    \begin{tabular}{|c|c|c|}
        \cline{2-3} \cline{3-3} 
        \multicolumn{1}{c|}{} & LS\_Cp & NN\_Cp\tabularnewline
        \hline 
        MAE (kW) & 113.196 & \textbf{15.341}\tabularnewline
        \hline 
        RMSE (kW) & 167.893 & \textbf{27.770}\tabularnewline
        \hline 
        MAPE (\%) & 37.65 & \textbf{3.63}\tabularnewline
        \hline 
        R2 & 0.8570 & \textbf{0.9961}\tabularnewline
        \hline 
    \end{tabular}
    \caption{Error metrics of the power coefficient models obtained using the least squares method and an artificial neural network.}
    \label{tab:metrics_CP}
\end{table}

Hereafter, standard and physics-informed neural networks with different target variables are compared. The hyperparameters were optimized with the Hyperband algorithm from the Keras Tuner library, within the space generated by the combinations of $\texttt{n\_layers}\in\{1,2,4\}$, $\texttt{n\_neurons}\in\{8,16,32,64,128\}$, $\texttt{learning\_rate}\in\{0.01,0.001,0.0001\}$ and $\texttt{activation\_function}\in\{ \text{`relu'}, \text{`tanh'} \}$. Optimal hyperparameters are shown in \autoref{tab:model-hyperparameters} for each case. The batch size was fixed to 128, mean absolute error as loss function, and a reduced learning rate when loss stopped improving. The dataset was split into $80\,\%$ for training and $20\,\%$ for testing, being 150 and 200 epochs enough for learning to stagnate, for NNs and PINNs, respectively. 

\begin{table}
\centering
\renewcommand{\arraystretch}{1.2}
\begin{tabular}{|c|c|c|c|}
\cline{2-4} \cline{3-4} \cline{4-4} 
\multicolumn{1}{c|}{} & NN\_Cp & NN\_T & NN\_P\tabularnewline
\hline 
\texttt{n\_layers} & 2 & 2 & 2\tabularnewline
\hline 
\texttt{n\_neurons} & 128 & 128 & 64\tabularnewline
\hline 
\texttt{learning\_rate} & 0.001 & 0.001 & 0.001\tabularnewline
\hline 
\texttt{activation\_function} & relu & relu & relu\tabularnewline
\hline 
\end{tabular}
\caption{Optimal hyperparamers of the artificial neural network models for the power coefficient, torque and power.}
\label{tab:model-hyperparameters}
\end{table}

The models with different output magnitudes were compared by computing the generated power and using that prediction to calculate the performance metrics. The predicted power of each model could be compared on the one hand with the measured power from the dataset, $P$, and on the other hand with the power calculated from the measured torque data through the mechanical relation $P_T=gT\omega$. \autoref{tab:model-metrics} shows regression metrics between the predicted power and the ground truth, considered as $P$ for the `Data' columns and $P_T$ for the `Phys' columns, for the test dataset. 

\begin{table}
\begin{adjustwidth}{-2.5cm}{-2.5cm}
\centering
\renewcommand{\arraystretch}{1.2} 
\resizebox{1.4\columnwidth}{!}{
\begin{tabular}{|c|c|c|c|c|c|c|c|c|c|c|c|c|}
\cline{2-13} \cline{3-13} \cline{4-13} \cline{5-13} \cline{6-13} \cline{7-13} \cline{8-13} \cline{9-13} \cline{10-13} \cline{11-13} \cline{12-13} \cline{13-13} 
\multicolumn{1}{c|}{} & \multicolumn{2}{c|}{NN\_Cp} & \multicolumn{2}{c|}{PINN\_Cp} & \multicolumn{2}{c|}{NN\_T} & \multicolumn{2}{c|}{PINN\_T} & \multicolumn{2}{c|}{NN\_P} & \multicolumn{2}{c|}{PINN\_P}\tabularnewline
\cline{2-13} \cline{3-13} \cline{4-13} \cline{5-13} \cline{6-13} \cline{7-13} \cline{8-13} \cline{9-13} \cline{10-13} \cline{11-13} \cline{12-13} \cline{13-13} 
\multicolumn{1}{c|}{} & Data & Phys & Data & Phys & Data & Phys & Data & Phys & Data & Phys & Data & Phys\tabularnewline
\hline 
MAE (kW) & \textbf{15.341} & 15.122 & 15.375 & 14.513 & 15.797 & \textbf{14.399} & 15.431 & 14.601 & \textbf{15.363} & 15.173 & 15.545 & 14.543\tabularnewline
\hline 
RMSE (kW) & \textbf{27.770} & 27.856 & 27.940 & 26.967 & 28.511 & \textbf{26.901} & 27.898 & 27.144 & \textbf{27.785} & 27.967 & 28.016 & 26.946\tabularnewline
\hline 
MAPE (\%) & \textbf{3.63} & 3.63 & 3.70 & 3.40 & 3.92 & \textbf{3.33} & 3.72 & 3.37 & \textbf{3.64} & 3.63 & 3.84 & 3.44\tabularnewline
\hline 
R2 & \textbf{0.9961} & 0.9962 & 0.9960 & 0.9964 & 0.9959 & \textbf{0.9964} & 0.9960 & 0.9963 & \textbf{0.9961} & 0.9961 & 0.9960 & 0.9964\tabularnewline
\hline 
\end{tabular}
} 
\caption{Error metrics between data/physical target and predicted power for different models.}
\label{tab:model-metrics}
\end{adjustwidth}
\end{table}

As a matter of fact, all models are good regressors of the power, having a MAE less than $16\,$kW and a MAPE smaller than $4\,\%$. The models reproduce well both $P$ and $P_T$, hence correctly describing both data and the physical equation governing the system simultaneously. A deeper analysis can be done when comparing the NN versus the PINN models for the same target variable. For $C_p$ and $P$ models, NNs reproduce data slightly better than PINNs, while PINNs reproduce the physical equation slightly better than standard NNs. This was expected because PINNs have a physical term in the loss function to respect the physics law. 

In contrast, the opposite happens for the $T$ models: PINN\_T predicts power data better than NN\_T, which in turn satisfy better the physical equation. This apparent contradiction is understood by noticing that the data term of the loss function of \autoref{eq:Ldata_T} contributes to the learning of the equation $P_T=gT\omega$, for the way in which the power is calculated through the predicted torque as an intermediate variable. If the model predicts $T$ correctly, it will correctly predict $P_T=gT\omega$. For its part, the physical term of \autoref{eq:Lphys_T} contributes to the learning of the measured $P$ data, because it leads to learning the variable $T_P=P/g\omega$, which gives measured data when converted to power.
\begin{table}\label{tab:models-evidential}
\renewcommand{\arraystretch}{1.2}
\centering
\begin{tabular}{|c|c|c|c|}
\cline{2-4} \cline{3-4} \cline{4-4} 
\multicolumn{1}{c|}{} & NN\_Cp\_evi & NN\_T\_evi & NN\_P\_evi\tabularnewline
\hline 
MAE (kW) & \textbf{15.080} & 15.692 & 15.241\tabularnewline
\hline 
RMSE (kW) & \textbf{27.188} & 28.236 & 27.606\tabularnewline
\hline 
MAPE (\%) & \textbf{3.58} & 3.91 & 3.61\tabularnewline
\hline 
R2 & \textbf{0.9963} & 0.9960 & 0.9961\tabularnewline
\hline 
MAU (kW) & 17.610 & 17.314 & 18.603\tabularnewline
\hline 
\end{tabular}
\caption{Error metrics between data and predicted power for different neural network models with an evidential output layer. MAU accounts for Mean Absolute Uncertainty.}
\label{tab:metrics_evidential}
\end{table}
It is also remarkable that NN models also satisfy the physical equation with high accuracy without imposing physical terms into the loss function. NN\_Cp and NN\_P are the best models for predicting power data and NN\_T to fulfill the physical \autoref{eq:Power-torque}, highlighted with bold in \autoref{tab:model-metrics}. 

\begin{figure}[t]
\begin{subfigure}{0.55\textwidth}
\includegraphics[width=1.0\linewidth]{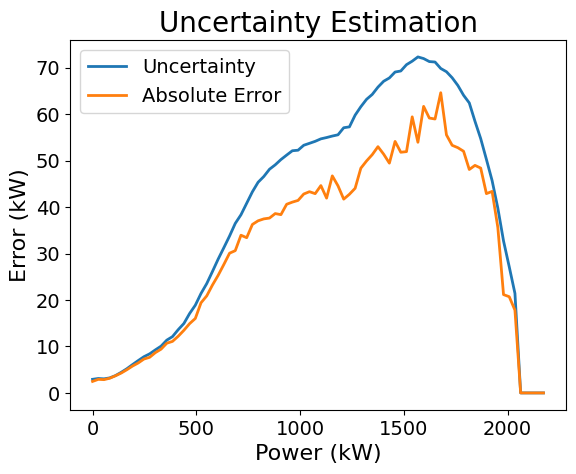} 
\caption{}
\label{fig:Error_vs_Power}
\end{subfigure}
\begin{subfigure}{0.45\textwidth}
\includegraphics[width=1.0\linewidth]{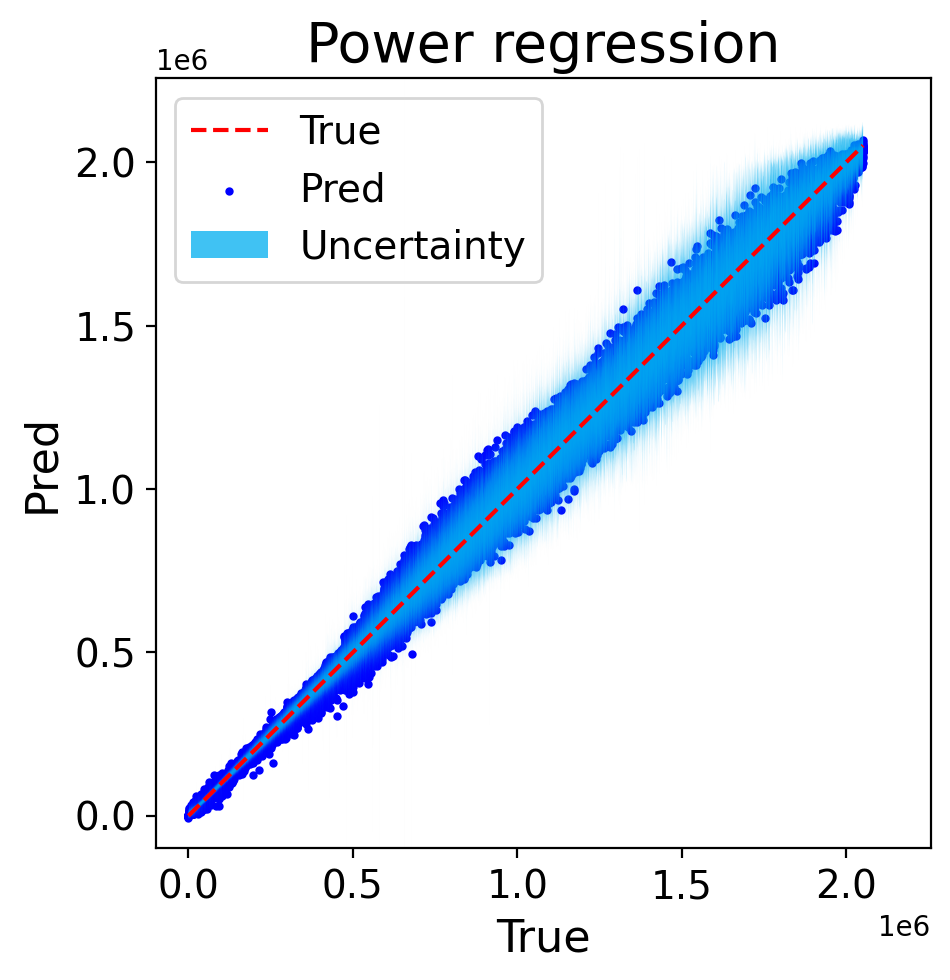}
\caption{}
\label{fig:fit_NN_P}
\end{subfigure}
\caption{(a) Absolute error and uncertainty as a function of power. (b) Predicted vs True values of power, along with the associated uncertainty ($\pm3\sigma$), for the test set.}
\end{figure}

\begin{figure}
     \centering
     \includegraphics[width=0.7\linewidth]{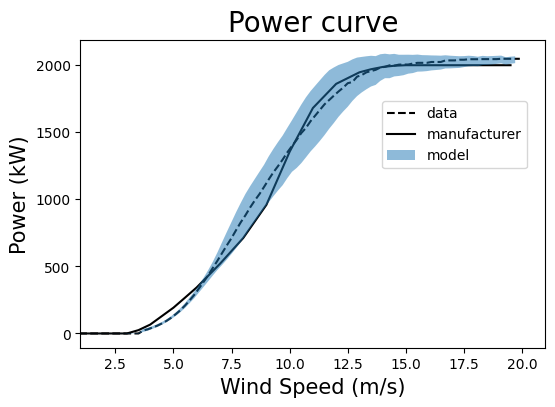}
     \caption{Comparison of power curve obtained from data, the model (uncertainty of $\pm3\sigma$) and the manufacturer.}
     \label{fig:Power_curve_model}
\end{figure}

As a last step, evidential deep learning models are built to predict power coefficient, torque, and power. Metrics between data and predicted power are shown in \autoref{tab:metrics_evidential}, all of them providing good results. Apart from the regression of the target magnitude, these models also provide the epistemic uncertainty (associated with the model) of each prediction. As can be noticed from the table, mean absolute uncertainty is slightly greater than mean absolute error, meaning that, on average, the prediction falls inside the 1$\sigma$ confidence interval. This happens in all power ranges, as can be observed in \autoref{fig:Error_vs_Power}, where absolute error and uncertainty are plotted against the power. This figure is associated with the model with $P$ as the target, but very similar curves are obtained for $C_p$ and $T$ models. Uncertainty is larger for intermediate and high power values, in agreement with \autoref{fig:fit_NN_P}, representing predicted versus true values for the test set. The power curve as a function of the wind speed obtained from the neural network model is depicted in \autoref{fig:Power_curve_model} together with the estimated curve from data and the one provided by the manufacturer \cite{MM82}. Uncertainty is represented in this figure with a confidence interval of 3$\sigma$, meaning that the real model falls inside that interval with a probability of $97.3\,\%$. Indeed, the curve estimated from data (dashed black line) is inside the confidence interval (blue) for all the wind speed range and the manufacturer curve (solid black line) for most of the range, slightly differing only at a wind velocity around 5\,m/s.

\section{Conclusions}\label{sec:conclusions}

In this work, we have designed and validated models for predicting the generated power of wind turbines as a function of wind velocity, pitch angle, and rotor angular velocity. They were trained using real historical data of 4 turbines from `La Haute Borne' wind farm. In the preprocessing, raw data were used to estimate the power curve as a function of the wind speed and $30\,\%$ of data were considered as anomalous points and thus discarded for the subsequent analysis.

The first approach was to fit the power coefficient with a parameterized empirical formula, but it showed poor results on the dataset under analysis, being unable to capture the non-linear relationship between the target magnitude and the wind velocity. The absence of a precise physical model describing the system and the availability of data makes it very suitable for applying data-driven approaches. Neural network models showed great performance when used for regression of the output variables like power coefficient, torque or power itself. Furthermore, the physical information provided by the state equation relating torque and power paved the way for the implementation of physics-informed neural networks, which were able to emulate both the data and the physical constraints adequately. In short, all neural network models yielded very good results, with a MAPE less than $4\,\%$ and a MAE below $16\,$kW of power. Finally, introducing an evidential output layer provided uncertainty quantification of the predictions, a desirable feature of any physical model. The predicted uncertainties (proved to be consistent with the absolute error) made possible the definition of confidence intervals in the power curve and a better agreement with the manufacturer specifications.

The deployed models could serve as regression-based anomaly detection methods, comparing the deviation of new data from the prediction of the model for a healthy state. Especially useful are models predicting power and an associate uncertainty, which quantifies the probability of new data to be normal or an anomaly. All the presented models are completely differentiable, so they could be used for building optimal pitch angle controllers, giving rise to optimal power generation at different wind speed regimes. Although this family of models seems promising, its robustness against different turbines would need further investigation. However, they could be easily transferred to different manufacturing by fine-tuning the parameters of the pre-trained models.

\section*{Acknowledments}
This work was partially funded by the Spanish Ministry of Economic Affairs and Digital Transformation (NextGenerationEU funds) within the project IA4TES MIA.2021.M04.0008. AG, MMS and JGR were also funded by FEDER/Junta de Andalucía (D3S project, P21.00247) and Spanish Ministry of Science (SINERGY, PID2021.125537NA.I00).

\section*{Author contributions}
\textbf{Alfonso Gijón:} Conceptualization, Methodology, Software, Validation, Investigation, Writing - Original Draft. \textbf{Ainhoa Pujana-Goitia:} Conceptualization, Writing - Review \& Editing, \textbf{Eugenio Perea:} Conceptualization \textbf{Miguel Molina-Solana:} Conceptualization, Writing - Review \& Editing, Supervision, Funding acquisition. \textbf{Juan Gómez-Romero:} Conceptualization, Writing - Review \& Editing, Supervision, Funding acquisition. 	

\appendix
\setcounter{figure}{0}

\section{Physics-Informed Neural Network Architecture}\label{sec:PINN_arhitecture}

\begin{figure}
    \centering
    \includegraphics[width=0.8\linewidth]{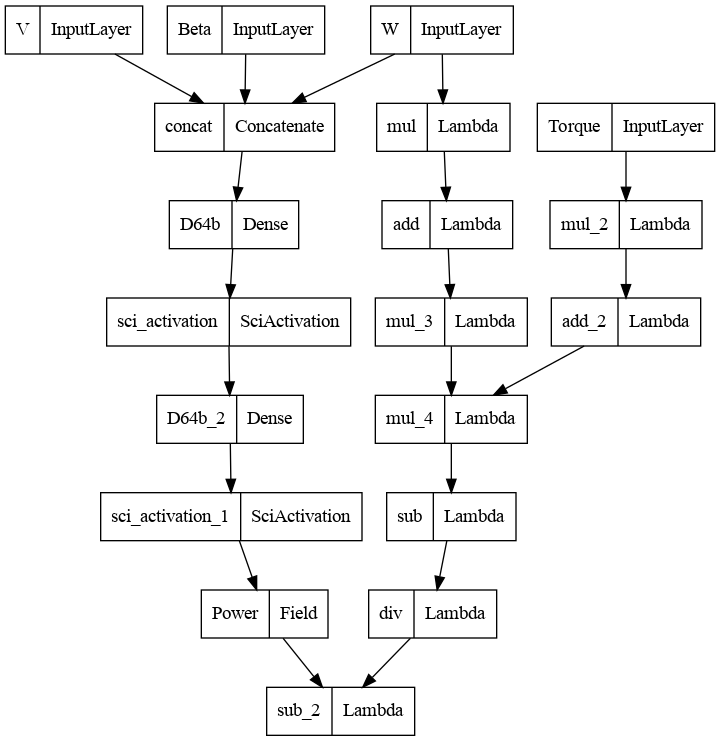}
    \caption{Model graph of the PINN predicting $P$ as a function of $(v,\beta,\omega)$ and respecting equation $P=gT\omega$.}
    \label{fig:PINN_architecture}
\end{figure}

As an example, the PINN architecture built for predicting the power while satisfying the physical equation $P=gT\omega$ is schematized in \autoref{fig:PINN_architecture}. The PINN receives $(v,\beta,\omega,T)$ as input, $(v,\beta,\omega)$ being the arguments of an artificial neural network with two hidden dense layers that output the power. Independently, the torque $T$ is combined with $\omega$ through lambda layers to compute $gT\omega$. Before the multiplication of $\omega$ and $T$, additional lambda layers were applied for unit conversion (input data was standardized to have zero mean and standard deviation equal to 1) before combining $\omega$ and $T$. Finally, the power in original units is transformed to standard units to be comparable with the output of the artificial neural network. This workflow allows the training of a PINN to minimize both terms of the loss function of Equations \ref{eq:Ldata_P} and \ref{eq:Lphys_P}. 


\bibliographystyle{elsarticle-num}
\biboptions{sort&compress} 
\bibliography{./references.bib}

\begin{thebibliography}{10}
\expandafter\ifx\csname url\endcsname\relax
  \def\url#1{\texttt{#1}}\fi
\expandafter\ifx\csname urlprefix\endcsname\relax\def\urlprefix{URL }\fi
\expandafter\ifx\csname href\endcsname\relax
  \def\href#1#2{#2} \def\path#1{#1}\fi

\bibitem{IRENA}
I.~R. E.~A. IRENA,
  \href{https://www.irena.org/Publications/2023/Jun/World-Energy-Transitions-Outlook-2023}{World
  energy transitions outlook 2023: 1.5°c pathway, volume 1, international
  renewable} (2023).
\newline\urlprefix\url{https://www.irena.org/Publications/2023/Jun/World-Energy-Transitions-Outlook-2023}

\bibitem{offshore_growup}
WindEurope,
  \href{https://windeurope.org/intelligence-platform/product/offshore-wind-energy-2022-mid-year-statistics/}{Offshore
  wind energy 2022 mid-year statistics} (2022).
\newline\urlprefix\url{https://windeurope.org/intelligence-platform/product/offshore-wind-energy-2022-mid-year-statistics/}

\bibitem{Wu2016}
J.~Wu, S.~Guo, J.~Li, D.~Zeng, Big data meet green challenges: Big data toward
  green applications, IEEE Systems Journal 10~(3) (2016) 888--900.
\newblock \href {https://doi.org/10.1109/JSYST.2016.2550530}
  {\path{doi:10.1109/JSYST.2016.2550530}}.

\bibitem{AKHAVANHEJAZI2018}
H.~Akhavan-Hejazi, H.~Mohsenian-Rad, Power systems big data analytics: An
  assessment of paradigm shift barriers and prospects, Energy Reports 4 (2018)
  91--100.
\newblock \href {https://doi.org/https://doi.org/10.1016/j.egyr.2017.11.002}
  {\path{doi:https://doi.org/10.1016/j.egyr.2017.11.002}}.

\bibitem{SAHAL2020}
R.~Sahal, J.~G. Breslin, M.~I. Ali, Big data and stream processing platforms
  for industry 4.0 requirements mapping for a predictive maintenance use case,
  Journal of Manufacturing Systems 54 (2020) 138--151.
\newblock \href {https://doi.org/https://doi.org/10.1016/j.jmsy.2019.11.004}
  {\path{doi:https://doi.org/10.1016/j.jmsy.2019.11.004}}.

\bibitem{Gomez2015}
C.~Q. Gómez, M.~A. Villegas, F.~P. García, D.~J. Pedregal, Big Data and Web
  Intelligence for Condition Monitoring, IGI-Global, 2015.
\newblock \href {https://doi.org/10.4018/978-1-4666-8505-5.ch008}
  {\path{doi:10.4018/978-1-4666-8505-5.ch008}}.

\bibitem{Canizo2017}
M.~Canizo, E.~Onieva, A.~Conde, S.~Charramendieta, S.~Trujillo, Real-time
  predictive maintenance for wind turbines using big data frameworks, in: 2017
  IEEE International Conference on Prognostics and Health Management (ICPHM),
  2017, pp. 70--77.
\newblock \href {https://doi.org/10.1109/ICPHM.2017.7998308}
  {\path{doi:10.1109/ICPHM.2017.7998308}}.

\bibitem{ML-WindTurbines-Review}
A.~Stetco, F.~Dinmohammadi, X.~Zhao, V.~Robu, D.~Flynn, M.~Barnes, J.~Keane,
  G.~Nenadic, Machine learning methods for wind turbine condition monitoring: A
  review, Renewable Energy 133 (2019) 620--635.
\newblock \href {https://doi.org/10.1016/j.renene.2018.10.047}
  {\path{doi:10.1016/j.renene.2018.10.047}}.

\bibitem{offshore_digitalization}
A.~Clifton, S.~Barber, A.~Bray, P.~Enevoldsen, J.~Fields, A.~M. Sempreviva,
  L.~Williams, J.~Quick, M.~Purdue, P.~Totaro, Y.~Ding, Grand challenges in the
  digitalisation of wind energy, Wind Energy Science 8~(6) (2023) 947–--974.
\newblock \href {https://doi.org/10.5194/wes-8-947-2023}
  {\path{doi:10.5194/wes-8-947-2023}}.

\bibitem{WT_modelling_challenges}
M.~Tan, Z.~Zhang, Wind turbine modeling with data-driven methods and radially
  uniform designs, IEEE Transactions on Industrial Informatics 12~(3) (2016)
  1261--1269.
\newblock \href {https://doi.org/10.1109/TII.2016.2532321}
  {\path{doi:10.1109/TII.2016.2532321}}.

\bibitem{AVENDANOVALENCIA2020106686}
L.~D. Avendaño-Valencia, E.~N. Chatzi, D.~Tcherniak, Gaussian process models
  for mitigation of operational variability in the structural health monitoring
  of wind turbines, Mechanical Systems and Signal Processing 142 (2020) 106686.
\newblock \href {https://doi.org/10.1016/j.ymssp.2020.106686}
  {\path{doi:10.1016/j.ymssp.2020.106686}}.

\bibitem{en16020861}
A.~Pujana, M.~Esteras, E.~Perea, E.~Maqueda, P.~Calvez, Hybrid-model-based
  digital twin of the drivetrain of a wind turbine and its application for
  failure synthetic data generation, Energies 16~(2) (2023).
\newblock \href {https://doi.org/10.3390/en16020861}
  {\path{doi:10.3390/en16020861}}.

\bibitem{WT_power_curve}
M.~Mehrjoo, M.~Jafari~Jozani, M.~Pawlak, Wind turbine power curve modeling for
  reliable power prediction using monotonic regression, Renewable Energy 147
  (08 2019).
\newblock \href {https://doi.org/10.1016/j.renene.2019.08.060}
  {\path{doi:10.1016/j.renene.2019.08.060}}.

\bibitem{10.1049/ip-gtd_19941215}
S.~Watson, L.~Landberg, J.~Halliday, Application of wind speed forecasting to
  the integration of wind energy into a large scale power system, IEE
  Proceedings - Generation, Transmission and Distribution 141 (1994) 357--362.
\newblock \href {https://doi.org/10.1049/ip-gtd_19941215}
  {\path{doi:10.1049/ip-gtd_19941215}}.

\bibitem{en14185967}
M.~Benbouzid, T.~Berghout, N.~Sarma, S.~Djurović, Y.~Wu, X.~Ma,
  \href{https://www.mdpi.com/1996-1073/14/18/5967}{Intelligent condition
  monitoring of wind power systems: State of the art review}, Energies 14~(18)
  (2021).
\newblock \href {https://doi.org/10.3390/en14185967}
  {\path{doi:10.3390/en14185967}}.
\newline\urlprefix\url{https://www.mdpi.com/1996-1073/14/18/5967}

\bibitem{APATA2020e00566}
O.~Apata, D.~Oyedokun,
  \href{https://www.sciencedirect.com/science/article/pii/S2468227620303045}{An
  overview of control techniques for wind turbine systems}, Scientific African
  10 (2020) e00566.
\newblock \href {https://doi.org/https://doi.org/10.1016/j.sciaf.2020.e00566}
  {\path{doi:https://doi.org/10.1016/j.sciaf.2020.e00566}}.
\newline\urlprefix\url{https://www.sciencedirect.com/science/article/pii/S2468227620303045}

\bibitem{Howland2019}
M.~F. Howland, J.~O. Dabiri, Wind farm modeling with interpretable
  physics-informed machine learning, Energies 12~(14) (2019).
\newblock \href {https://doi.org/10.3390/en12142716}
  {\path{doi:10.3390/en12142716}}.

\bibitem{RAISSI2019}
M.~Raissi, P.~Perdikaris, G.~Karniadakis, Physics-informed neural networks: A
  deep learning framework for solving forward and inverse problems involving
  nonlinear partial differential equations, Journal of Computational Physics
  378 (2019) 686--707.
\newblock \href {https://doi.org/10.1016/j.jcp.2018.10.045}
  {\path{doi:10.1016/j.jcp.2018.10.045}}.

\bibitem{FMata2023}
F.~{Fernández de la Mata}, A.~Gijón, M.~Molina-Solana, J.~Gómez-Romero,
  Physics-informed neural networks for data-driven simulation: Advantages,
  limitations, and opportunities, Physica A: Statistical Mechanics and its
  Applications 610 (2023) 128415.
\newblock \href {https://doi.org/10.1016/j.physa.2022.128415}
  {\path{doi:10.1016/j.physa.2022.128415}}.

\bibitem{Kuik2016}
G.~A.~M. van Kuik, J.~Peinke, R.~Nijssen, D.~Lekou, J.~Mann, J.~N. S{\o}rensen,
  C.~Ferreira, J.~W. van Wingerden, D.~Schlipf, P.~Gebraad, H.~Polinder,
  A.~Abrahamsen, G.~J.~W. van Bussel, J.~D. S{\o}rensen, P.~Tavner, C.~L.
  Bottasso, M.~Muskulus, D.~Matha, H.~J. Lindeboom, S.~Degraer, O.~Kramer,
  S.~Lehnhoff, M.~Sonnenschein, P.~E. S{\o}rensen, R.~W. K\"unneke, P.~E.
  Morthorst, K.~Skytte, Long-term research challenges in wind energy - a
  research agenda by the european academy of wind energy, Wind Energy Science
  1~(1) (2016) 1--39.
\newblock \href {https://doi.org/10.5194/wes-1-1-2016}
  {\path{doi:10.5194/wes-1-1-2016}}.

\bibitem{evidential2020}
A.~Amini, W.~Schwarting, A.~Soleimany, D.~Rus, Deep evidential regression, in:
  Proceedings of the 34th International Conference on Neural Information
  Processing Systems, NIPS'20, Curran Associates Inc., Red Hook, NY, USA, 2020.

\bibitem{10.1063/5.0091980}
N.~Zehtabiyan-Rezaie, A.~Iosifidis, M.~Abkar, {Data-driven fluid mechanics of
  wind farms: A review}, Journal of Renewable and Sustainable Energy 14~(3)
  (2022) 032703.
\newblock \href {https://doi.org/10.1063/5.0091980}
  {\path{doi:10.1063/5.0091980}}.

\bibitem{MM82}
{Repower MM82 - Manufacturers and Turbines - Online access - The Wind Power},
  \url{https://www.thewindpower.net/turbine_en_16_repower_mm82.php},
  {Accessed}: 09-05-2023.

\bibitem{doi:https://doi.org/10.1002/9781119992714.ch3}
Aerodynamics of Horizontal Axis Wind Turbines, John Wiley \& Sons, Ltd, 2011,
  Ch.~3, pp. 39--136.
\newblock \href {https://doi.org/https://doi.org/10.1002/9781119992714.ch3}
  {\path{doi:https://doi.org/10.1002/9781119992714.ch3}}.

\bibitem{IEC61400-12-1}
I.~C. Secretary, \href{https://webstore.iec.ch/publication/68499}{Wind
  turbines—part 12-1: Power performance, measurements of electricity
  producing wind turbines {Part} 12-1: {Overview}} (2022).
\newline\urlprefix\url{https://webstore.iec.ch/publication/68499}

\bibitem{PAGNINI2015}
L.~C. Pagnini, M.~Burlando, M.~P. Repetto, Experimental power curve of
  small-size wind turbines in turbulent urban environment, Applied Energy 154
  (2015) 112--121.
\newblock \href
  {https://doi.org/https://doi.org/10.1016/j.apenergy.2015.04.117}
  {\path{doi:https://doi.org/10.1016/j.apenergy.2015.04.117}}.

\bibitem{YAN2019}
J.~Yan, H.~Zhang, Y.~Liu, S.~Han, L.~Li, Uncertainty estimation for wind energy
  conversion by probabilistic wind turbine power curve modelling, Applied
  Energy 239 (2019) 1356--1370.
\newblock \href
  {https://doi.org/https://doi.org/10.1016/j.apenergy.2019.01.180}
  {\path{doi:https://doi.org/10.1016/j.apenergy.2019.01.180}}.

\bibitem{Jamdade2015}
P.~G. Jamdade, S.~G. Jamdade, Modelling and simulation of mechanical torque
  developed by wind turbine generator excited with different wind speed
  profiles, in: V.~Vijay, S.~K. Yadav, B.~Adhikari, H.~Seshadri, D.~K. Fulwani
  (Eds.), Systems Thinking Approach for Social Problems, Springer India, New
  Delhi, 2015, pp. 267--278.
\newblock \href {https://doi.org/10.1007/978-81-322-2141-8_23}
  {\path{doi:10.1007/978-81-322-2141-8_23}}.

\bibitem{Carpintero2020}
M.~Carpintero‐Renteria, D.~Santos‐Martin, A.~Lent, C.~Ramos, Wind turbine
  power coefficient models based on neural networks and polynomial fitting, IET
  renewable power generation. 14~(11) (2020-08).

\bibitem{cp_equations}
O.~C. Castillo, V.~R. Andrade, J.~J.~R. Rivas, R.~O. González, Comparison of
  power coefficients in wind turbines considering the tip speed ratio and blade
  pitch angle, Energies 16~(6) (2023).
\newblock \href {https://doi.org/10.3390/en16062774}
  {\path{doi:10.3390/en16062774}}.

\bibitem{WANG2023}
Y.~Wang, X.~Duan, R.~Zou, F.~Zhang, Y.~Li, Q.~Hu, A novel data-driven deep
  learning approach for wind turbine power curve modeling, Energy 270 (2023)
  126908.
\newblock \href {https://doi.org/10.1016/j.energy.2023.126908}
  {\path{doi:10.1016/j.energy.2023.126908}}.

\bibitem{Zhang2008}
J.~Zhang, M.~Cheng, Z.~Chen, X.~Fu, Pitch angle control for variable speed wind
  turbines, in: 2008 Third International Conference on Electric Utility
  Deregulation and Restructuring and Power Technologies, 2008, pp. 2691--2696.
\newblock \href {https://doi.org/10.1109/DRPT.2008.4523867}
  {\path{doi:10.1109/DRPT.2008.4523867}}.

\bibitem{Abdul-Ruhman2020}
M.~S. Abdul-Ruhman, M.~N. Hawas, Optimal pitch angle control for wind turbine
  using intelligent controller, IOP Conference Series: Materials Science and
  Engineering 745~(1) (2020) 012017.
\newblock \href {https://doi.org/10.1088/1757-899X/745/1/012017}
  {\path{doi:10.1088/1757-899X/745/1/012017}}.

\bibitem{SierraGarcia2022-2}
J.~E. Sierra-Garcia, M.~Santos, Deep learning and fuzzy logic to implement a
  hybrid wind turbine pitch control, Neural Computing and Applications 34~(13)
  (2022) 10503--10517.
\newblock \href {https://doi.org/10.1007/s00521-021-06323-w}
  {\path{doi:10.1007/s00521-021-06323-w}}.

\bibitem{Karami2023}
A.~Karami-Mollaee, O.~Barambones, Pitch control of wind turbine blades using
  fractional particle swarm optimization, Axioms 12~(1) (2023).
\newblock \href {https://doi.org/10.3390/axioms12010025}
  {\path{doi:10.3390/axioms12010025}}.

\bibitem{Bilendo2021}
F.~Bilendo, H.~Badihi, N.~Lu, P.~Cambron, B.~Jiang, An intelligent data-driven
  machine learning approach for fault detection of wind turbines, in: 2021 6th
  International Conference on Power and Renewable Energy (ICPRE), 2021, pp.
  444--449.
\newblock \href {https://doi.org/10.1109/ICPRE52634.2021.9635340}
  {\path{doi:10.1109/ICPRE52634.2021.9635340}}.

\bibitem{SUN2021}
L.~Sun, F.~You, Machine learning and data-driven techniques for the control of
  smart power generation systems: An uncertainty handling perspective,
  Engineering 7~(9) (2021) 1239--1247.
\newblock \href {https://doi.org/https://doi.org/10.1016/j.eng.2021.04.020}
  {\path{doi:https://doi.org/10.1016/j.eng.2021.04.020}}.

\bibitem{MANOBEL2018}
B.~Manobel, F.~Sehnke, J.~A. Lazzús, I.~Salfate, M.~Felder, S.~Montecinos,
  Wind turbine power curve modeling based on gaussian processes and artificial
  neural networks, Renewable Energy 125 (2018) 1015--1020.
\newblock \href {https://doi.org/10.1016/j.renene.2018.02.081}
  {\path{doi:10.1016/j.renene.2018.02.081}}.

\bibitem{LI2022}
T.~Li, X.~Liu, Z.~Lin, R.~Morrison, Ensemble offshore wind turbine power curve
  modelling – an integration of isolation forest, fast radial basis function
  neural network, and metaheuristic algorithm, Energy 239 (2022) 122340.
\newblock \href {https://doi.org/10.1016/j.energy.2021.122340}
  {\path{doi:10.1016/j.energy.2021.122340}}.

\bibitem{SierraGarcia2021}
J.~E. Sierra-García, M.~Santos, Neural networks and reinforcement learning in
  wind turbine control, Revista Iberoamericana de Automática e Informática
  industrial 18~(4) (2021) 327–335.
\newblock \href {https://doi.org/10.4995/riai.2021.16111}
  {\path{doi:10.4995/riai.2021.16111}}.

\bibitem{SierraGarcia2022}
J.~E. Sierra-Garcia, M.~Santos, R.~Pandit, Wind turbine pitch reinforcement
  learning control improved by pid regulator and learning observer, Engineering
  Applications of Artificial Intelligence 111 (2022) 104769.
\newblock \href
  {https://doi.org/https://doi.org/10.1016/j.engappai.2022.104769}
  {\path{doi:https://doi.org/10.1016/j.engappai.2022.104769}}.

\bibitem{Coquelet2022}
M.~Coquelet, L.~Bricteux, M.~Moens, P.~Chatelain, A reinforcement-learning
  approach for individual pitch control, Wind Energy 25~(8) (2022) 1343--1362.
\newblock \href {https://doi.org/https://doi.org/10.1002/we.2734}
  {\path{doi:https://doi.org/10.1002/we.2734}}.

\bibitem{XIE2023}
J.~Xie, H.~Dong, X.~Zhao, Data-driven torque and pitch control of wind turbines
  via reinforcement learning, Renewable Energy 215 (2023) 118893.
\newblock \href {https://doi.org/10.1016/j.renene.2023.06.014}
  {\path{doi:10.1016/j.renene.2023.06.014}}.

\bibitem{LI2022-2}
X.~Li, W.~Zhang, Physics-informed deep learning model in wind turbine response
  prediction, Renewable Energy 185 (2022) 932--944.
\newblock \href {https://doi.org/10.1016/j.renene.2021.12.058}
  {\path{doi:10.1016/j.renene.2021.12.058}}.

\bibitem{Jacques2023}
J.~H. Mclean, M.~R. Jones, B.~J. O’Connell, E.~Maguire, T.~J. Rogers,
  Physically meaningful uncertainty quantification in probabilistic wind
  turbine power curve models as a damage-sensitive feature, Structural Health
  Monitoring 0~(0) (2023) 14759217231155379.
\newblock \href {https://doi.org/10.1177/14759217231155379}
  {\path{doi:10.1177/14759217231155379}}.

\bibitem{Aerts2023}
F.~Aerts, L.~Lanzilao, J.~Meyers, Bayesian uncertainty quantification framework
  for wake model calibration and validation with historical wind farm power
  data, Wind Energy 26~(8) (2023) 786--802.
\newblock \href {https://doi.org/10.1002/we.2841} {\path{doi:10.1002/we.2841}}.

\bibitem{Marepally2022}
K.~Marepally, Y.~S. Jung, J.~Baeder, G.~Vijayakumar, Uncertainty quantification
  of wind turbine airfoil aerodynamics with geometric uncertainty, Journal of
  Physics: Conference Series 2265~(4) (2022) 042041.
\newblock \href {https://doi.org/10.1088/1742-6596/2265/4/042041}
  {\path{doi:10.1088/1742-6596/2265/4/042041}}.

\bibitem{Pascal2022}
P.~Richter, J.~Wolters, M.~Frank, Uncertainty quantification of offshore wind
  farms using monte carlo and sparse grid, Energy Sources, Part B: Economics,
  Planning, and Policy 17~(1) (2022) 2000520.
\newblock \href {https://doi.org/10.1080/15567249.2021.2000520}
  {\path{doi:10.1080/15567249.2021.2000520}}.

\bibitem{Perr-Sauer2021}
J.~Perr-Sauer, M.~Optis, J.~M. Fields, N.~Bodini, J.~C. Lee, A.~Todd,
  E.~Simley, R.~Hammond, C.~Phillips, M.~Lunacek, T.~Kemper, L.~Williams,
  A.~Craig, N.~Agarwal, S.~Sheng, J.~Meissner, Openoa: An open-source codebase
  for operational analysis of wind farms, Journal of Open Source Software
  6~(58) (2021) 2171.
\newblock \href {https://doi.org/10.21105/joss.02171}
  {\path{doi:10.21105/joss.02171}}.

\bibitem{tensorflow}
M.~Abadi, A.~Agarwal, P.~Barham, E.~Brevdo, Z.~Chen, C.~Citro, G.~S. Corrado,
  A.~Davis, J.~Dean, M.~Devin, S.~Ghemawat, I.~Goodfellow, A.~Harp, G.~Irving,
  M.~Isard, Y.~Jia, R.~Jozefowicz, L.~Kaiser, M.~Kudlur, J.~Levenberg,
  D.~Man\'{e}, R.~Monga, S.~Moore, D.~Murray, C.~Olah, M.~Schuster, J.~Shlens,
  B.~Steiner, I.~Sutskever, K.~Talwar, P.~Tucker, V.~Vanhoucke, V.~Vasudevan,
  F.~Vi\'{e}gas, O.~Vinyals, P.~Warden, M.~Wattenberg, M.~Wicke, Y.~Yu,
  X.~Zheng, \href{https://www.tensorflow.org/}{{TensorFlow}: Large-Scale
  Machine Learning on Heterogeneous Systems} (2015).

\bibitem{hyperband}
L.~Li, K.~Jamieson, G.~DeSalvo, A.~Rostamizadeh, A.~Talwalkar, Hyperband: A
  novel bandit-based approach to hyperparameter optimization, J. Mach. Learn.
  Res. 18~(1) (2017) 6765--6816.

\bibitem{keras}
T.~O'Malley, E.~Bursztein, J.~Long, F.~Chollet, H.~Jin, L.~Invernizzi, et~al.,
  Kerastuner, \url{https://github.com/keras-team/keras-tuner} (2019).

\bibitem{sciann}
E.~Haghighat, R.~Juanes, Sciann: A keras/tensorflow wrapper for scientific
  computations and physics-informed deep learning using artificial neural
  networks, Computer Methods in Applied Mechanics and Engineering 373 (2021)
  113552.

\bibitem{LaHauteBorne1}
R.~Mandzhieva, R.~Subhankulova, Data-driven applications for wind energy
  analysis and prediction: The case of “la haute borne” wind farm, Digital
  Chemical Engineering 4 (2022) 100048.
\newblock \href {https://doi.org/https://doi.org/10.1016/j.dche.2022.100048}
  {\path{doi:https://doi.org/10.1016/j.dche.2022.100048}}.

\bibitem{LaHauteBorne2}
N.~Effenberger, N.~Ludwig, A collection and categorization of open-source wind
  and wind power datasets, Wind Energy 25~(10) (2022) 1659--1683.
\newblock \href {https://doi.org/https://doi.org/10.1002/we.2766}
  {\path{doi:https://doi.org/10.1002/we.2766}}.

\bibitem{Minh2018}
D.~Minh-Thang, J.~Berthaut-Gerentes,
  \href{https://windeurope.org/workshops/wp-content/uploads/files/Tech18a/files/posters/PO024.pdf}{An
  estimation of key metrics from scada data - a third-party's view}, in: Poster
  presented at Analysis of Operating Wind Farms conference, 2018.

\end{thebibliography}

\end{document}